\pdfoutput=1

\documentclass[11pt]{article}

\usepackage[]{EMNLP2023}

\usepackage{times}
\usepackage{latexsym}

\usepackage{paralist}
\usepackage[T1]{fontenc}

\usepackage[utf8]{inputenc}

\usepackage{microtype}

\usepackage{inconsolata}

\usepackage{amsmath,amssymb,amsfonts}
\usepackage{url,enumitem}
\usepackage{booktabs}
\usepackage{xspace}
\usepackage{graphicx}
\usepackage{subcaption}
\usepackage{comment}
\usepackage[normalem]{ulem}
\usepackage{mathtools}
\usepackage{svg}
\usepackage{algorithm,algorithmicx,algpseudocode}
\usepackage{dsfont}

\usepackage{wrapfig}
\usepackage{tabularx}
\usepackage{booktabs}
\usepackage{dsfont}
\usepackage{multirow}
\usepackage{array}
\usepackage{caption}
\usepackage{listings}
\usepackage{adjustbox}

\usepackage{listings}
\newcommand{\Sref}[1]{\S\ref{#1}}

\newcommand{\red}[1]{\textcolor{red}{[#1]}}
\newcommand{\orange}[1]{\textcolor{orange}{[#1]}}
\newcommand{\green}[1]{\textcolor{green}{[#1]}}

\newcommand{\toolname}[1]{\textsc{Facts\&Evidence}}
\newcommand{\dolly}[1]{\textsc{dolly}}

\newcommand{\ignore}[1]{}

\let\orgautoref\autoref

\renewcommand{\autoref}[1]{\def\equationautorefname{Eq.}\orgautoref{#1}}
\newcommand{\Tref}[1]{Table~\ref{#1}}
\newcommand{\Fref}[1]{Figure~\ref{#1}}

\title{
\toolname{}: An Interactive Tool for Transparent Fine-Grained  Factual Verification of Machine-Generated Text
}

\author{
  Varich Boonsanong$^{\spadesuit}$ \quad \quad Vidhisha Balachandran$^\diamondsuit$ \quad
  \textbf{Xiaochuang Han}$^\spadesuit$ \\ \textbf{Shangbin Feng}$^{\spadesuit}$ \quad \quad \textbf{Lucy Lu Wang}$^{\spadesuit}$ \quad \quad \textbf{Yulia Tsvetkov}$^{\spadesuit}$\\
  $^\spadesuit$University of Washington \qquad $^\diamondsuit$Microsoft Research \qquad \\
        {\tt \{varicb, xhan77, shangbin, lucylw, yuliats\}@cs.washington.edu} \\ {\tt{vidhishab@microsoft.com}} 
}

\begin{document}
\maketitle

\begin{abstract} 
With the widespread consumption of AI-generated content, there has been an increased focus on developing automated tools to verify the factual accuracy of such content. However, prior research and tools developed for fact verification treat it as a binary classification or a linear regression problem.
Although this is a useful mechanism as part of automatic guardrails in systems, we argue that such tools lack transparency in the prediction reasoning and diversity in source evidence to provide a trustworthy user experience.


We develop \toolname{}---an interactive and transparent tool for user-driven verification of complex text. The tool facilitates the intricate decision-making involved in fact-verification, presenting its users a breakdown of complex input texts to visualize the credibility of individual claims along with an explanation of model decisions and attribution to multiple, diverse evidence sources.
\toolname{} aims to empower consumers of machine-generated text 
and give them agency to understand, verify, selectively trust and use such text.
\end{abstract}

\begin{figure*}[t]
\centering
\includegraphics[width=1.0\textwidth]{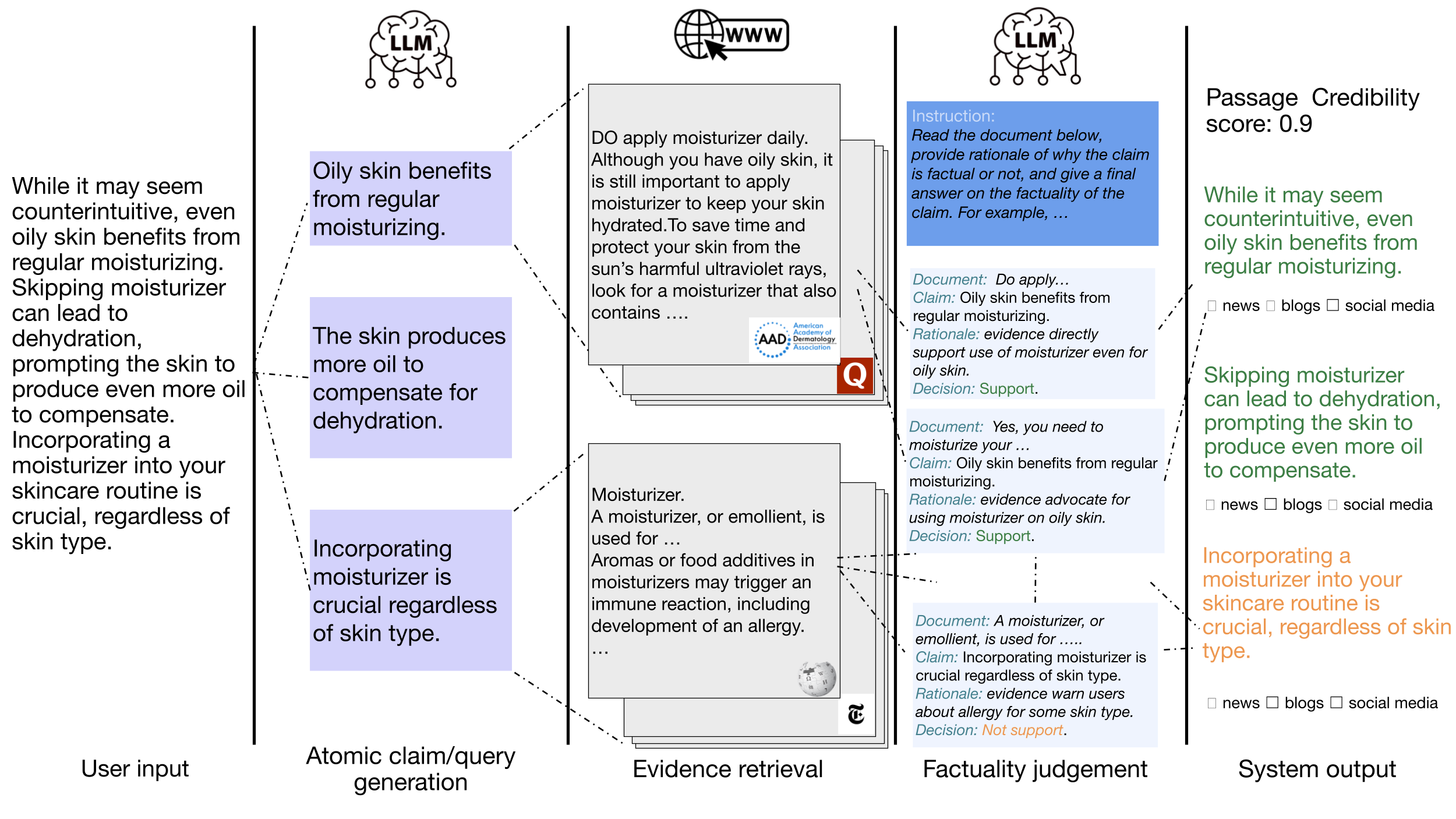}
\caption{
The pipeline figure of \toolname{}. The user input for verification goes through atomic claim and query generation, evidence retrieval, and factuality judgement processes. The system output is aggregated from the judgements over atomic claims, providing users with sources of evidence and an overall credibility score. 
}
\label{fig:main_fig}
\end{figure*}

\section{Introduction}

In our time of proliferating use and adoption of AI-generated text and AI-assisted writing, the need for robust and transparent fact verification tools to help users verify the accuracy of machine-generated content has never been more pressing. Errors and hallucinations generated by AI models are already translating into real-world problems through errors in generated code introducing unintended bugs \citep{chgpt-code-error}, nonexistent books and references propagating misinformation \citep{chgpt-ref-error1, chgpt-ref-error2, chgpt-ref-error3}, and incorrect medical advice causing health complications \citep{chgpt-med-error}, which cost stakeholders time, money, and in some cases leading to physical harm \citep{chgpt-harm-error, JOHNSON2022101753}. Although fact-verification of machine-generated texts has been studied extensively in recent research \citep{Manakul2023SelfCheckGPTZB, gao-etal-2023-rarr, Huang2023ASO, Li2024TheDA}, this work has not resulted in practical tools that are accessible to users, easy to use, and  transparent  \citep{Do2024FacilitatingHC}. As such, developing fact-verification tools \emph{aimed at informing users of the veracity of AI-generated content} is necessary to enable users to understand, verify, and trust such content.

Recent research has developed methods for automatically detecting factual errors and hallucinations through self-checking with Large Language Models (LLMs) \citep{Manakul2023SelfCheckGPTZB}, uncertainty calibration \citep{farquhar2024detecting, zhang2023enhancing, feng2024don} or fine-tuning specific factuality classifiers \citep{mishra2024fine}, and tools for visualizing such errors for users \citep{fatahi-bayat-etal-2023-fleek, Krishna2024GenAuditFF}. The primary intent of this research has been the binary classification (factual vs.~non-factual) of complex text to serve as an intrinsic guardrail in AI systems. As a result, their decisions are categorical, often not explainable, and can be noisy \citep{mishra2024fine}.


On the other hand, real-world human fact-checking is significantly more complex, involving verification against diverse sources with varying levels of reliability and leanings \citep{augenstein2019multifc, dias2020researching, glockner-etal-2022-missing}. Evidence for verifying facts is traced through various sources, and claims are verified against each source. The authenticity and reliability of each source or news outlet is considered when aggregating and deciding the veracity of each claim. Such fine-grained views into each claim per source provide a detailed understanding of the correctness of long, complex texts, allowing consumers of the text to assess which specific claims to trust and use.

In this work, we take a step towards addressing this gap.
We present \toolname{}---a user-driven, transparent and interactive fact-verification tool to empower consumers of AI-generated text to understand the factual accuracy of the text. \toolname{} takes any long, complex text from a user as input and presents a breakdown of the text into individual claims. Each claim is searched on the web to identify multiple, diverse evidence items and verify against each of them. Ultimately, decisions from multiple pieces of evidence and claims are aggregated to compute a credibility score for the entire text. In addition to aggregate decisions, \toolname{} presents the per-claim, per-evidence decision, along with a model-generated rationale explaining the decision and a source tag indicating the source category of the evidence. Further, \toolname{} is interactive, allowing users to include or exclude any source type from the verification process based on personal preferences and specific needs. 

We evaluate the reliability of decisions produced by \toolname{} using the FAVA dataset \cite{mishra2024fine} and show that the predicted scores are competitive with strong baselines, outperforming previous systems by $\sim$ 40 F1 points, and establishing a high-quality system in claim verification of open-ended generation tasks. In summary, our contribution is \toolname{}---a high-quality, fine-grained, and interactive tool providing transparency and access to real-time information in the fact-verification process which can inform and empower users, giving them the ability to inspect and validate the veracity of each part of the content they use.\footnote{Demo URL: \\ \url{https://factsandevidence.cs.washington.edu/}\\
Video URL: \\
\url{https://www.youtube.com/watch?v=_MaAI2H7c3w}}


\section{\toolname{} -- User Interface}
\label{sec:ui}
This section presents 
the user interface of \toolname{} when it is used to inspect the veracity of machine-generated texts.

\subsection{Landing Page:} 
\begin{figure}[t!]
\includegraphics[width=0.5\textwidth]{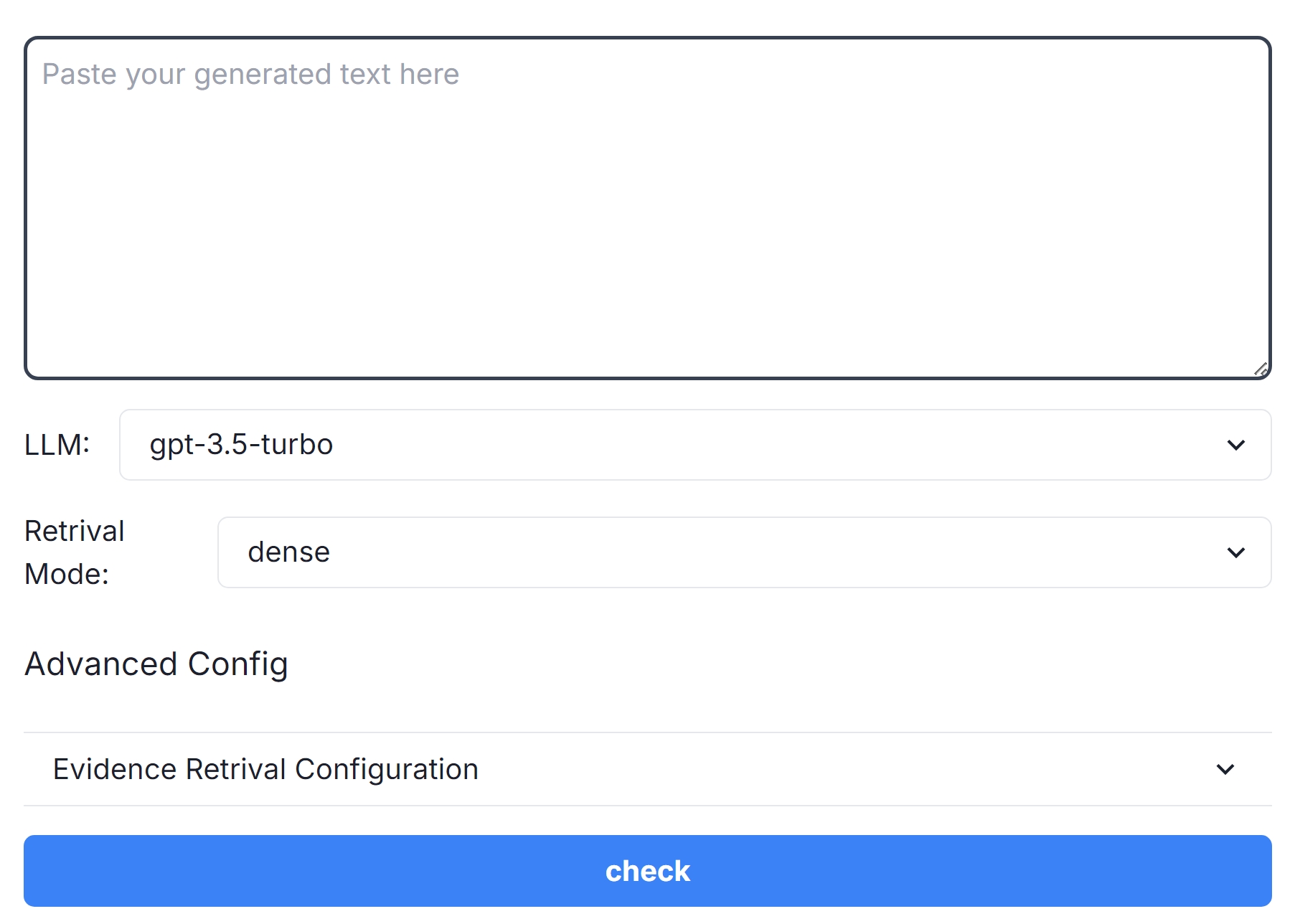}\caption{\toolname{} Upload Panel} \label{fig:input_panel}
    \vspace{-0.4cm}
\end{figure}

\paragraph{Upload Panel:} The first screen (ref: \Fref{fig:input_panel}) presents a panel for the user to upload the text they want to verify. Additionally, the user can configure different parameters associated with their search: (i) LLM: choice of $2$ backend LLMs (\texttt{gpt-3.5-turbo-0301} and \texttt{Llama3-8B}) used for verifying claims; (ii) Retrieval Mode: choice of dense (embedding-based) or sparse (keyword-based) retrieval for extracting relevant evidence from web search text; (iii) Evidence Configuration: choice of top-$k$ web results to parse, top-$k$ relevant passages within each search result, and context size of each passage. For each of these parameters, default values are set based on internal evaluations for best claim verification performance and user experience as outlined in \Sref{sec:expts}. Once the text is provided and the user is satisfied with the parameters, they can submit their request to verify the text. 

\paragraph{Readme:} The home screen also includes a README
that explains the objectives of the tool and a brief explanation of how the system conducts fact verification behind the scenes to provide full transparency to users. It describes how a user can use the tool and what level of interaction they can expect from the tool. In addition, it includes an explanation for each configurable parameter and possible choices available to the user.

\subsection{Fact Visualization Page:} Once the user submits their text, we run our fact verification engine on our back-end and present a detailed view of the final results to the user. 

\begin{figure}[t!]
\centering
\includegraphics[width=0.45\textwidth]{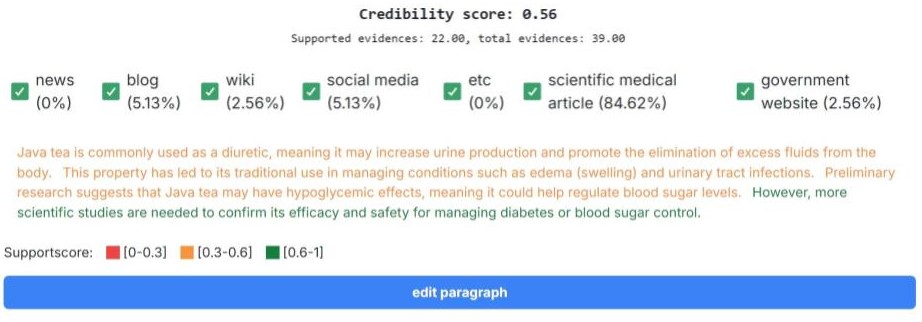}\caption{Credibility Panel of \toolname{}
} \label{fig:credibility_panel}
\includegraphics[width=0.45\textwidth]{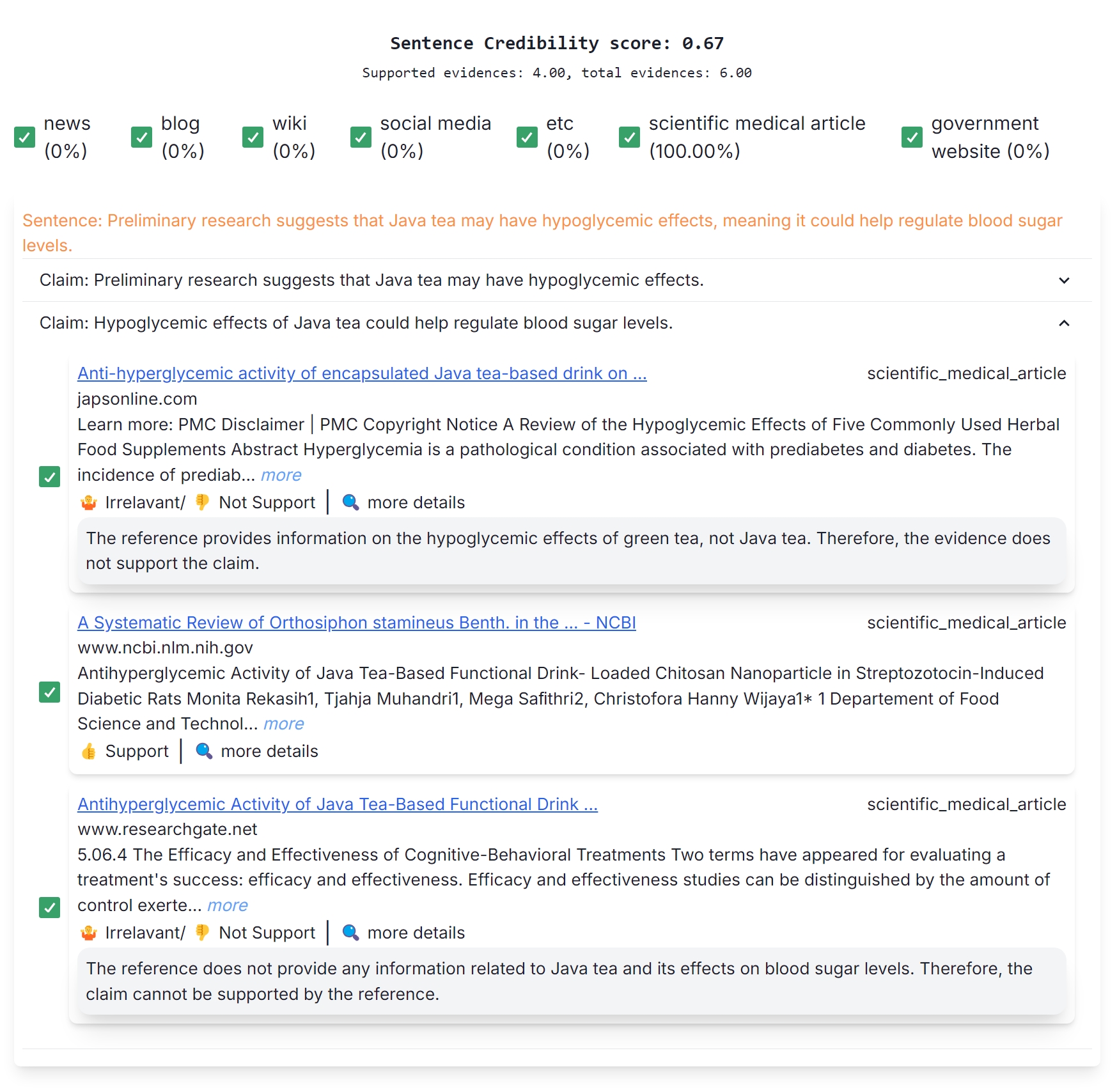}\caption{Evidence Panel of \toolname{}
} \label{fig:evidence_panel}
    \vspace{-0.4cm}
\end{figure}

\paragraph{Credibility Panel:} First the credibility panel (ref: \Fref{fig:credibility_panel}) presents a view of the input text with the credibility of individual sentences highlighted. A \emph{credibility score} is computed for each sentence in the text. Each sentence is visualized with a color coded with \red{red} for low credibility [0-0.3), \orange{orange} for medium credibility [0.3-0.6) and \green{green} for high credibility [0.6-1]. Finally, a credibility score for the entire text is also presented based on aggregated sentence level scores.

\paragraph{Evidence Panel:} 
For each sentence, individual atomic claims are displayed in an evidence panel (ref: \Fref{fig:evidence_panel}) with an expanded view showing multiple evidences from the Web used for prediction, a claim-evidence level factuality prediction (supported/not supported/irrelevant), and a rationale which explains the prediction in free-form text. Each evidence has a source type tag associated with it. Source types indicate the general type of web page the evidence was extracted from: news, scientific article, blogs, etc. 

The Evidence Panel gives users the option to include specific evidence documents or sources of evidence based on their preferences. 
Checkboxes can be selected to include or exclude each evidence document and recompute the credibility score, allowing users to inspect how specific evidence impacts the summary score. Users are also provided source-type check boxes at a global level to include or exclude certain sources they do not wish to use. For example, when verifying healthcare-related text users may choose to exclude blogs and social media posts as evidence and only focus on scientific articles or government websites for information. Credibility scores are automatically updated based on the source and evidence selections made by the users. This fine-grained view and user control into the evidences and source types allows user agency in customizing the fact verification process for different domains and inputs.

\section{\toolname{} -- Behind the Scenes}
\label{sec:backend}
This section details the back-end framework for verifying facts in model-generated texts. For providing a detailed view of facts, evidence items,  and their factual accuracy, the user text is broken down to individual claims, and for each claim, multiple, diverse supporting evidence items are retrieved from the web. A factual accuracy prediction and explanation for each claim is produced using an LLM and the resulting details are displayed to the user as outlined in \Sref{sec:ui}. \Fref{fig:main_fig} presents an outline of our process which we describe in detail below.

\paragraph{Atomic Claim Generation}
To break down the user text into individual atomic claims, we prompt an LLM to break a paragraph into its constituent sentences and break each sentence into a list of atomic claims. To accommodate context-length limits of different LLMs, we syntactically break long paragraphs into smaller ones and assume that each paragraph's context is self-contained. 
We prompt an LLM in a few-shot setting to decontextualize and generate a claim that can be used for web search. Atomic claims often assume context of the larger text, including pronouns and other references that can adversely impact our search for relevant evidence. For example, the claim `Headaches are a common side effect of this treatment' does not include relevant information about which treatment the claim refers to. Incorporating decontextualization instructions addresses this by specifically instructing the LLM to include any additional context required to make the claim sufficiently detailed for querying the Web.

\paragraph{Evidence Retrieval}
Given a list of atomic claims and their queries, we conduct a Google web search using the \texttt{Serper}\footnote{\url{https://serper.dev/}} service and scrape the top-$N$ websites as evidence using the \texttt{readability}\footnote{\url{https://github.com/mozilla/readability}} library hosted on \texttt{Cloudflare Worker}\footnote{\url{https://workers.cloudflare.com/}}. We identify the most relevant text within each source evidence using a text retrieval model (\texttt{jina-embeddings-v3}).
The evidence context is constructed by taking $M$ sentences before and after each relevant evidence sentence. In the case where the user prefers multiple passages within a document, this process can be repeated for top-$K$ similar sentences instead of just the best match sentence. In the end, the evidence contexts for an atomic claim can be multiple paragraphs of highly relevant text chunks from diverse sources of evidence.

\paragraph{Source categorization}
To provide users with the option of including or excluding an entire source or category of evidence, we classify each evidence source into a set of predefined categories. We extract the hostname of the website and ask an LLM in a zero-shot fashion to select one of the closest related categories among \{news, blog, wiki, social media, scientific/medical article, government website, other\}. These categories were chosen to cover a wide range of evidence sources and can be expanded further in the future.

\begin{figure*}[!t]
    \begin{center}
\includegraphics[width=0.9\linewidth]{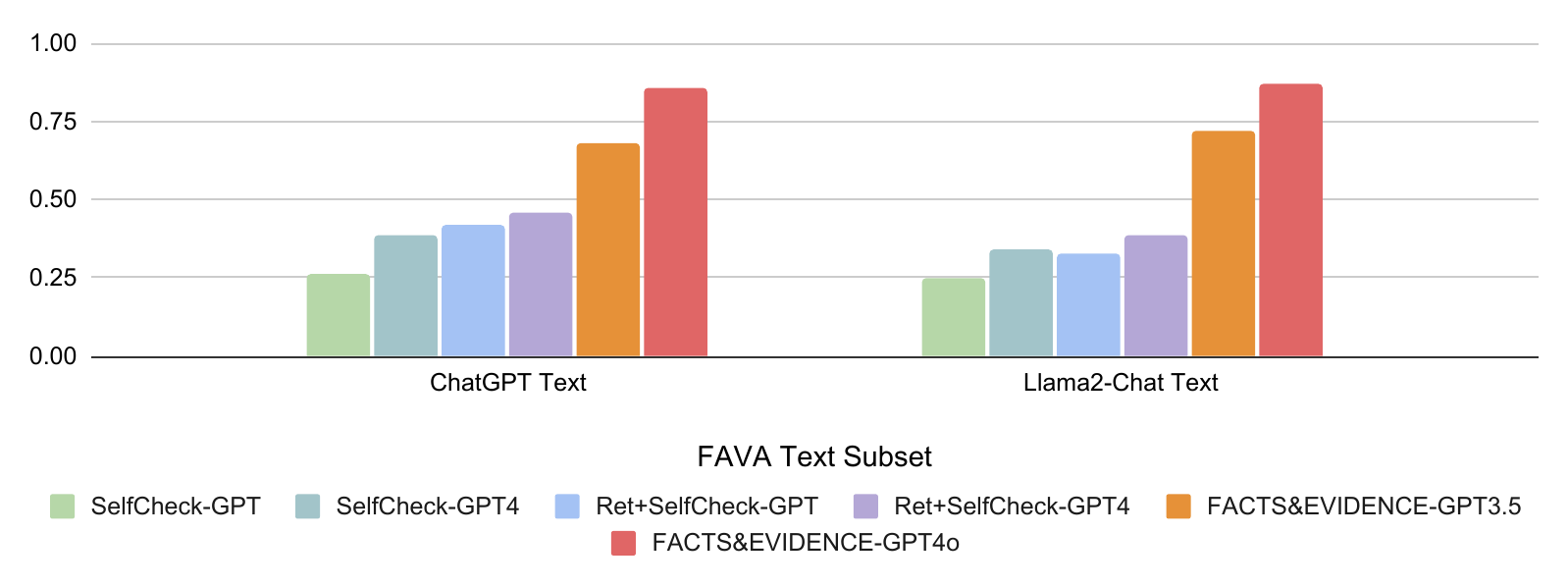}\vspace{-3mm}
    \end{center}
    \caption{Binary F1 Results on factual error detection. \textbf{\toolname{} improves error prediction accuracy by $\sim$44 points on average across the two subsets.}  }\label{fig:res_main}\vspace{-4.5mm}
\end{figure*}

\paragraph{Factuality Judgement}
We use chain-of-thought style prompting \citep{NEURIPS2022_9d560961} by giving the retrieved evidence and atomic claim to LLM to determine whether the evidence can support the atomic claim or not. We extract the model's decision and rationale to present in the UI so that users can validate the decision and decide whether or not it should be included in the aggregate factuality assessment. We weigh each piece of evidence equally and calculate a sentence-level and passage-level credibility score as the \emph{percentage of evidence that support the claims} in a sentence/passage.
\[
    \text{credibility score} = \frac{\text{\# support evidence}}{\text{\# total evidence}}
\]


To promote reusability, we create our back-end API to comply with open API standards. Any developer can call any intermediate steps of our tools through this API interface. Additionally, all prompts used to query LLMs in the various steps are included in Appendix \Sref{app:prompts}.

\section{Experiments}
\label{sec:expts}
\paragraph{Evaluation:} 
We evaluate our predictions using the FavaBench dataset \citep{mishra2024fine}. FavaBench includes human factuality annotations at the span level for approximately 1,000 model-generated responses from two popular LLMs (ChatGPT and Llama2-Chat) for 200 diverse information-seeking and knowledge-intensive queries.
For ChatGPT and Llama2-Chat, 59.8\% and 65.5\% of the respective responses include at least one hallucination, respectively. We obtain sentence-level labels by associating any sentence with a factually incorrect span with an incorrect sentence label. We run the FavaBench inputs through our system using \texttt{gpt-3.5-turbo-0301} for all LLM calls, and for each sentence we predict Not-Factual if \emph{any} of the evidences does not support the claims (credibility score < $t$)\footnote{$t=0.3$ considering 3 evidences}.
Following \citet{mishra2024fine}, we report sentence-level F1 against human-annotated judgments on the ChatGPT and Llama2-Chat subsets of the data separately.



\begin{table}[t!]
\centering
\footnotesize
\addtolength{\tabcolsep}{-1.1pt}  
\begin{tabular}{p{3.2cm}p{1.4cm}p{1.9cm}}\toprule
Setting & ChatGPT Text & Llama2-Chat Text \\
\midrule
No:Ev 1, Ctxt W 15 & 0.23 & 0.25  \\
No:Ev 1, Ctxt W 30 & 0.23 & 0.22 \\
No:Ev 3, Ctxt W 15 & 0.68 & 0.72 \\
No:Ev 3, Ctxt W 30 & 0.64 & 0.67 \\
No Atomic Claim Gen & 0.40 & 0.43\\
\bottomrule
\end{tabular}
    \vspace{-0.1cm}
    \caption{Results of Ablation Study 
    }
    \vspace{-0.32cm}
\label{tab:ablation_results}
\end{table}

\paragraph{Baselines:}
We compare predictions from our tool against multiple strong baselines: (i) SelfCheck-ChatGPT \citep{Manakul2023SelfCheckGPTZB} prompts ChatGPT (\texttt{gpt-3.5-turbo-0301}) to reflect and identify factual errors in the generated text; (ii) SelfCheck-ChatGPT4 (\texttt{gpt-4}) follows the same setup with GPT4 instead; (iii) Ret+SelfCheck-GPT follows the setup from \citet{mishra2024fine} and \citet{min2023factscore} to prompt ChatGPT to identify factual errors in the text given the retrieved document from a static retrieval system (Contriever-MSMarco with Wikipedia documents) to augment the original prompt; (iv) Ret+SelfCheck-ChatGPT4 follows the same setup with GPT4 instead.

\paragraph{Overall Results:} 
In 
\Fref{fig:res_main},
we present the sentence-level binary F1 results on FavaBench. Our results show that \toolname{} performs better than prior work, showing that the tool predictions are high quality and reliable. In particular, \toolname{} outperforms all baselines that directly use LLMs with or without evidence to identify factual errors in complex text (SelfCheck-GPT, Ret+SelfCheck-GPT, SelfCheck-GPT4, Ret+SelfCheck-GPT4).
Specifically, compared to previous works, which rely on static knowledge sources and single evidence to verify facts, our tool breaks complex text to individual claims, dynamically retrieves diverse evidence from the web, and uses multiple evidence for verifying each fact, which improves the quality of our predictions.
\begin{table}[t!]
\centering
\footnotesize
\addtolength{\tabcolsep}{-1.8pt}  
\begin{tabular}{p{1.8cm}p{0.037\textwidth}p{0.037\textwidth}}\toprule
Retriever & \multicolumn{1}{c}{ChatGPT Text} & \multicolumn{1}{c}{Llama2-Chat Text} \\
\midrule
BM25 & 0.46 & 0.47  \\
Distilbert & 0.59 & 0.72 \\
Snowflake & 0.84 & 0.86 \\
Jina & 0.84 & 0.87 \\
\bottomrule
\end{tabular}
    \vspace{-0.2cm}
    \caption{Analysis of varying evidence retrievers}
    \vspace{-0.4cm}
\label{tab:retriever}
\end{table}

\begin{table*}[h!]\footnotesize
    \centering
\renewcommand{\arraystretch}{1.5} 
    \begin{tabular}{|p{4cm}p{8cm}p{2cm}|}
        \hline
        \multicolumn{3}{|p{14cm}|}{Sentence: Java tea is commonly used as a diuretic, meaning it may increase urine production. The property has led to its traditional use in managing conditions such as edema (swelling) and UTIs. } \\ \hline
        \textbf{Error Type} & \textbf{Example} & \textbf{\% Occurrence} \\ \hline
        Incorrect Atomic Claim & It is traditionally used to manage conditions such as edema.  & 26.7\%  \\ \hline
        Incorrect Evidence Retrieval & There’s a lot of different types of unflavored black tea, and the type and where it’s grown makes a big difference. & 53.3\% \\ \hline
        Incorrect Factuality Judgment & Not Factual prediction for `Java tea is commonly used as a diuretic'   & 20.0\% \\ \hline
    \end{tabular}
    \caption{Error types and their occurrences. Predominant failures in \toolname{} arises in evidence retrival.}
    \vspace{-0.25cm}\label{tab:qualitativeAnalysis}
\end{table*}

\noindent \textbf{Ablation Study:} In \Tref{tab:ablation_results}, we present an ablation study that varies the number of evidences and the size of the context window of each evidence to study the effect of evidences on performance. Our results indicate that our performance increases with an increased number of evidence, validating our hypothesis that complex fact checking requires multiple diverse evidences for high-quality predictions. However, the increase in the context window does not significantly increase the F1 score. This enables us to build a more efficient demo by using a smaller number of tokens as evidence, reducing the load time, and improving the user experience. Additionally, to validate the importance and quality of the atomic claim generator, we ablate the pipeline by removing the claim generation. Our results show significant drop in performance confirming that breaking complex claims to atomic claims is essential for fact verification.

In addition to evidence representation, we also conducted a study on the retriever model in \Tref{tab:retriever}. The Jina's embedding model (\texttt{jinaai/jina-embeddings-v3}) has the highest F1 score, followed by Snowflake embedding model (\texttt{snowflake-arctic-embed-l}), and other industry standard retrieval systems such as BM25 and Distillbert (\texttt{msmarco-distilbert-dot-v5}). \\ [-3mm]

\noindent\textbf{Qualitative Analysis}
We also performed a qualitative error analysis on 30 randomly sampled incorrect predictions from \toolname{} to identify parts of the pipeline that led to incorrect predictions \Tref{tab:qualitativeAnalysis}. Our analysis indicates that most of the errors in the pipeline are due to evidence retrieval failures. Such failures often stem from search engine failing to find relevant documents or website scraper failures due to security blocks or parsing errors. These failures are external to the system and could potentially be mitigated by increasing the number of evidence at a higher cost. Other errors in the pipeline were due to incorrect atomic claims (improper decontextualization, complex claims) or incorrect factuality judgements. Future work could explore custom finetuned models for the individual steps to mitigate these errors.

\section{Related Work}
\noindent\textbf{Factuality and Hallucination in LLMs:}
Recently, a focus has been on addressing the problem of model hallucinations \citep{kumar-etal-2023-language, mallen-etal-2023-trust, min2023factscore}. 
Previous research has studied the variety of errors and hallucinations produced by different models and characterized them using taxonomies for summarization \citep{pagnoni2021understanding}, text simplification \citep{devaraj-etal-2022-evaluating}, dialogue generation \citep{10.1162/tacl_a_00529, gupta2021dialfact}, and open-ended generation \citep{mishra2024fine}. Automated methods were developed to identify and flag model errors and hallucinations \citep{min2023factscore, mishra2024fine, gao-etal-2023-rarr, Yuksekgonul2023AttentionSA}. 
Although hallucination detection methods are primarily used as safety guardrails in systems, early tools have been developed to detect and present users with factual errors in model-generated text \citep{Krishna2024GenAuditFF, fatahi-bayat-etal-2023-fleek}. While these tools primarily present single evidence and final model decisions, \toolname{} provides a transparent and fine-grained view of the factual accuracy of model-generated text with an interactive interface for consumers with diverse evidences, detailed credibility scores, and explanations. \\ [-3mm]



\noindent\textbf{Fact-Verification:}
In parallel, there has been active research on automated fact verification of human written text and claims as well \citep{Walter2019FactCheckingAM, Guo2021ASO, 10.1145/3485127}. 
Various methods, tools, and systems have been built for automated fact checking of human-written claims \citep{thorne-vlachos-2018-automated, ijcai2021p619}. However, these techniques and systems have not been tested for LM-generated text. Although the aim of \toolname{} is to support the verification and understanding of facts in the text generated by a model and has been explicitly evaluated for this setting, it can be directly adapted and used to verify written human claims.

\section{Conclusion} 
We presented \toolname{}---an interactive tool for user-driven fact verification of AI-generated text. The tool enables consumers of outputs from various LLMs to visualize and understand individual claims and ground the claims in diverse evidence from the Web.
We evaluate the decisions made by the \toolname{} system and show that our system outperforms strong baselines in the claim verification task. \toolname{} promotes transparency and user agency in the verification process of facts and aims to empower consumers to understand fine-grained claims and appropriately use text generated by AI systems.

\section*{Limitations}
Though our goal is to provide a transparent user facing fact-verification tool for consumers of model generated text, our system internally relies on models for multiple steps in the pipeline. Inherently, there can arise concerns of accuracy and reliability of the pipeline. Through evaluations we showed that our system produces high-quality predictions for most of the cases. But as with any automated system, errors can arise. We hope that by designing and interactive user interface that exposes all the intermediate steps of the pipeline, our users will be able to view and judge our system decisions. Our goal is to maintain the final agency with the user directly. In future, we would like to include measures of confidence in model generations and predictions and present that information to the user as well to support them in their decision making.

Another limitation of our current system is efficiency. To improve the reliability, accuracy and transparency of our system, we introduced multiple sequential steps in the pipeline with multiple LLM calls to external APIs. While this enabled us to inspect and improve each step, we had to trade off some speed. Our demo currently takes 30s-1min for passages of 5-7 sentences. We plan to focus on improving the efficiency of the system in the future by introducing local small LMs that are finetuned for individual steps.

\section*{Acknowledgments}
This research was developed with funding from the Defense Advanced Research Projects Agency's (DARPA) SciFy program (Agreement No.~HR00112520300). The views expressed are those of the author and do not reflect the official policy or position of the Department of Defense or the U.S.~Government.
We also gratefully acknowledge support from  the University of Washington Population Health Initiative, the National Science Foundation (NSF) CAREER Grant No.~IIS2142739, and gift funding from Google and the Allen Institute for AI.

\bibliography{my_cites, custom}
\bibliographystyle{acl_natbib}

\clearpage

\appendix
\section{LLM Prompts}
\label{app:prompts}

Here we share the LLM prompts and instructions used at different stages of the pipeline:


\begin{table*}[t]
\resizebox{1\textwidth}{!}{
\begin{tabularx}{1.2\textwidth}{X}
\toprule[0.1em]
\texttt{
Given a sentence, your task is to separate a given sentence into series of short claims which are standalone fully self-contained sentences. You will be given a paragraph with sentences marked and a particular target sentence within it. You should produce atomic claims for that sentences in a way that is consistent with the paragraph. Each claim should be theirs own bullet points.  Each claim should be decontextualized such that it should be understandable without the previous context. Use the context of the paragraph or sentence to replace any references or claims. You don't need to repeat the questions, just directly give me claims.}\\
\texttt{Here are some examples of sentences and their atomic claims:}\\
\texttt{Paragraph: S1: <example1 sentence1> S2: <example1 sentence2> S3: <example1 sentence3> S4: <example1 sentence4>}\\
\texttt{Break the following sentence into atomic claims: S2: <example sentence2>}\\
\texttt{Claim\_1: <example claim1>}\\
\texttt{Claim\_2: <example claim2>}\\
\texttt{Claim\_3: <example claim3>}\\
\texttt{Claim\_4: <example claim4>}\\ \\
\texttt{Paragraph: S1: <example2 sentence1> S2: <example2 sentence2> S3: <example2 sentence3> S4: <example2 sentence4> S5: <example2 sentence5>}\\
\texttt{Break the following sentence into atomic claims:
S1: <example sentence1>}\\
\texttt{Claim\_1: <example claim1>}\\
\texttt{Claim\_2: <example claim2>}\\
\texttt{Claim\_3: <example claim3>}\\
\texttt{Claim\_4: <example claim4>}\\
\texttt{Claim\_5: <example claim5>}\\
\texttt{Claim\_6: <example claim6>}\\
\\
\texttt{Paragraph: <input paragraph>}\\
\texttt{Break the following sentence into atomic claims: S8: <input sentence>}\\
\midrule
\end{tabularx}
}
\caption{\textbf{Prompt for Atomic Claim Generation.} }
\label{table:prompt_remove_instr}
\end{table*}

\begin{table*}[t]
\resizebox{1\textwidth}{!}{
\begin{tabularx}
{1.2\textwidth}{X}
\toprule
\texttt{Given both the hostname of the website and the list of categories. Choose the best category that will best describe this hostname. Only give the final valid category from the list without any explanation.
Categories: [
        "news",
        "blog",
        "wiki",
        "social\_media",
        "etc",
        "scientific\_medical\_article",
        "government\_website",
    ] Hostname: <input hostname>}\\
\midrule
\end{tabularx}
}
\caption{\textbf{Prompt for Classifying the source type.} }
\label{table:prompt_remove_instr}
\end{table*}


\begin{table*}[t]
\resizebox{1\textwidth}{!}{
\begin{tabularx}
{1.2\textwidth}{X}
\toprule
\texttt{
Given the claim, its context, and its evidence. You must find whether the evidence can contradict the claim in this context. The context is the original text that the claim is derived from. Critically evaluate whether the evidence shows any sign of contradiction or doesn't directly support the claim.}\\
\texttt{Here are some tips on what may cause evidence contradiction:}\\
\texttt{1. if the claim is subjective, the evidence may not be supporting the claim and that is not contradiction.}\\
\texttt{2. if the evidence is inconsistent or contradictory, it is not supporting the claim and that is not contradiction.}\\
\texttt{3. if the evidence is talking about a different topic or not related to the claim, that is not contradiction.}\\
\texttt{4. if the context have contradictory information with the evidence, but not with the specific claim that are focusing on, it is not contradiction.}\\
\texttt{Here are examples of how you should respond.}\\
\texttt{Passage: <example context>}\\
\texttt{Claim: <example claim>}\\
\texttt{Evidence: <example evidence>}\\
\texttt{Rationale: <output rationale>}\\
\texttt{Final Verdict: <output prediction>}\\
\texttt{Passage: <example context>}\\
\texttt{Claim: <example claim>}\\
\texttt{Evidence: <example evidence>}\\
\texttt{Rationale: <output rationale>}\\
\texttt{Final Verdict: yes.}\\
\texttt{Let's think through this step by step.}\\
\texttt{Passage: <context>}\\
\texttt{Claim: <claim>}\\
\texttt{Evidence: <evidence>}\\
\texttt{Give your answer in the following format.}\\
\texttt{Rationale: <rationale>}\\
\texttt{Final Verdict: <yes for evidence agrees with the claim, no otherwise>}\\
\texttt{give a summarized answer as the last line with either say yes for support or no for not supported
}\\
\midrule
\end{tabularx}
}
\caption{\textbf{Prompt for judging factuality.} }
\label{table:prompt_factuality judgement}
\end{table*}

\end{document}